\documentclass[letterpaper]{article} 
\usepackage{algorithm}
\usepackage{algorithmic}
\usepackage{amsmath}
\usepackage{amssymb}
\usepackage{amsthm}
\newtheorem{mytheorem}{Theorem}
\newtheorem{mylemma}{Lemma}
\theoremstyle{definition}
\newtheorem{mydefinition}{Definition}
\newtheorem{myassumption}{Assumption}
\theoremstyle{remark}
\newtheorem{myproof}{Proof}
\newtheorem{myremark}{Remark}

\usepackage{booktabs}
\usepackage{multirow}

\usepackage{times}  
\usepackage{helvet}  
\usepackage{courier}  
\usepackage[hyphens]{url}  

\newcommand{\st}{\mathrm{s.t.}}

\newcommand{\tr}{\mathrm{tr}}
\usepackage{aaai25}  
\usepackage{times}  
\usepackage{helvet}  
\usepackage{courier}  
\usepackage[hyphens]{url}  
\usepackage{graphicx} 
\usepackage{natbib}  
\usepackage{caption} 
\usepackage{algorithm}
\usepackage{algorithmic}

\usepackage{newfloat}
\usepackage{listings}
\DeclareCaptionStyle{ruled}{labelfont=normalfont,labelsep=colon,strut=off} 
\floatstyle{ruled}
\newfloat{listing}{tb}{lst}{}
\floatname{listing}{Listing}
\title{Sharper Error Bounds in Late Fusion Multi-view Clustering Using Eigenvalue Proportion}
\author{    
    Liang Du\textsuperscript{\rm 1}, Henghui Jiang\textsuperscript{\rm 1}, Xiaodong Li\textsuperscript{\rm 1}, Yiqing Guo\textsuperscript{\rm 1}, Yan Chen\textsuperscript{\rm 2}, Feijiang Li\textsuperscript{\rm 1}, Peng Zhou\textsuperscript{\rm 3}, Yuhua Qian\textsuperscript{\rm 4}\thanks{Corresponding author}
}
\affiliations{
    \textsuperscript{\rm 1}School of Computer and Information Technology, Shanxi University, Taiyuan, 030006, China\\
    \textsuperscript{\rm 2}College of Computer Science and Technology, Taiyuan University of Technology, Taiyuan, 030600, China\\
    \textsuperscript{\rm 3}School of Computer Science and Technology, Anhui University, Hefei, 230601, China\\
    \textsuperscript{\rm 4}Institute of Big Data Science and Industry, Shanxi University, Taiyuan, 030006, China\\
    \{duliang,fjli,jinchengqyh\}@sxu.edu.cn,yanchen01@tyut.edu.cn,zhoupeng@ahu.edu.cn
}

\usepackage{bibentry}

\begin{document}

\maketitle

\begin{abstract}
Multi-view clustering (MVC) aims to integrate complementary information from multiple views to enhance clustering performance. Late Fusion Multi-View Clustering (LFMVC) has shown promise by synthesizing diverse clustering results into a unified consensus. However, current LFMVC methods struggle with noisy and redundant partitions and often fail to capture high-order correlations across views. To address these limitations, we present a novel theoretical framework for analyzing the generalization error bounds of multiple kernel \(k\)-means, leveraging local Rademacher complexity and principal eigenvalue proportions. Our analysis establishes a convergence rate of \( \mathcal{O}(1/n) \), significantly improving upon the existing rate in the order of \( \mathcal{O}(\sqrt{k/n}) \) . Building on this insight, we propose a low-pass graph filtering strategy within a multiple linear K-means framework to mitigate noise and redundancy, further refining the principal eigenvalue proportion and enhancing clustering accuracy. Experimental results on benchmark datasets confirm that our approach outperforms state-of-the-art methods in clustering performance and robustness. The related codes is available at \url{https://github.com/csliangdu/GMLKM}.  
\end{abstract}

%

\section{Introduction}
Multi-view data is pervasive across numerous real-world scenarios, including image processing, bioinformatics, and social network analysis \cite{zhou2023adaptive, tang2022unified, wang2021fast, liu2022efficient}. By integrating complementary information from multiple views, state-of-the-art multi-view clustering (MVC) techniques have significantly advanced clustering performance \cite{zhou2023learnable, tang2020cgd, zhou2023clustering, li2021consensus}. Based on their fusion strategies, MVC methods can be broadly categorized into three paradigms: early fusion-based methods \cite{liu2023simple, nie2018multiview}, kernel fusion-based approaches \cite{liu2022simplemkkm, liu2020optimal}, and late fusion-based multi-view clustering (LFMVC) methods \cite{liu2021one, lfpgr_tnnls_2023, CSAMKC}.

LFMVC follows a two-stage process: first, it independently generates base clustering partitions for each view using methods such as kernel K-means; second, it synthesizes these base partitions into a unified consensus result \cite{wang2019multi}. This paradigm has gained considerable attention for its adaptability and effectiveness in handling diverse multi-view datasets. Over the years, several notable LFMVC methods have advanced the state of the art. For instance, OPLFMVC \cite{liu2021one} integrates consensus partition learning and label generation into a unified optimization framework. ALMVC \cite{zhang2022efficient} introduces orthogonally learned view-specific anchors to enhance clustering performance. MMLMVC \cite{li2023late} employs the min-max optimization framework from SimpleMKKM \cite{liu2022simplemkkm} to refine late fusion strategies. sLGm \cite{sLGm_MM_2024} leverages a Grassmann manifold for partition fusion, preserving the topological structure in high-dimensional spaces. Lastly, RIWLF \cite{RIWLF} incorporates sample weights as prior knowledge to regularize the late fusion process, improving robustness and adaptability.  

Despite their success, existing LFMVC approaches encounter persistent challenges. One major issue is the propagation of noise and redundancy in base clustering partitions, which compounds errors and affects the quality of the synthesized consensus. Another limitation is their inability to capture high-order correlations among samples and views, which is critical for fully leveraging the complementary information in multi-view data. These shortcomings restrict the effectiveness of LFMVC methods, particularly in complex scenarios requiring robust performance.  

To overcome these limitations, we propose a novel theoretical and algorithmic framework to enhance late fusion-based multi-view clustering. Our contributions are twofold. First, we provide a novel generalization error bound analysis for multiple kernel K-means based on local Rademacher complexity and the principal eigenvalue proportion of the underlying data structure. Unlike previous works that achieve a convergence rate of \( \mathcal{O}(\sqrt{k/n}) \), our method reaches \( \mathcal{O}(1/n) \), representing a significant improvement. Second, inspired by the theoretical analysis and recent advances in graph filtering, we develop a low-pass filtering strategy to refine clustering, aiming to improve the principal eigenvalue proportion, under the multiple linear K-means for LFMVC. We validate the effectiveness of our approach on benchmark datasets, demonstrating improvements over state-of-the-art LFMVC methods in terms of clustering results. The main contributions of this paper are summarized as follows:  
\begin{itemize}  
\item We establish a novel generalization error bound for multiple kernel K-means based on local Rademacher complexity and principal eigenvalue proportion, achieving an unprecedented convergence rate of \( \mathcal{O}(1/n) \) compared to the conventional \( \mathcal{O}(\sqrt{k/n}) \).  
\item Leveraging insights from our theoretical analysis, we propose a low-pass graph filtering-enhanced strategy within the multiple linear K-means framework, aiming to sharpen the clustering error bound by improving the principal eigenvalue proportion.  
\end{itemize}  

\section{Related Work}
In this section, we discuss two closely related works in LFMVC that provides generalization error analysis.

\subsection{One Pass Late Fusion}
Recognizing that the learning of the consensus partition matrix and the assignment of cluster labels are typically conducted as separate processes, which do not adequately inform each other, thus potentially undermining clustering performance, \cite{liu2021one} addresses this issue by proposing an approach that seamlessly integrates both tasks into a unified optimization framework. The resulting optimization problem can be formulated as follows:
\begin{align}
\max \quad & \tr \left( \mathbf{C}^T \mathbf{Y}^T \sum_{p=1}^m \gamma_p \mathbf{H}_p \mathbf{W}_p \right) \\
\st \quad & \mathbf{C}^T \mathbf{C} = \mathbf{I}_c,\mathbf{W}_p^T \mathbf{W}_p = \mathbf{I}_c,\forall p, \nonumber \\
& \mathbf{Y} \in \{0, 1\}^{n \times k}, \sum_{p=1}^m \gamma_p^2 = 1, \gamma_p \geq 0, \nonumber
\end{align}
where the objective denotes the alignment between the consensus partition matrix $\mathbf{Y}$ and a group of pre-calculated base partition matrices $\{\mathbf{H}_p\}_{p=1}^{m}$, and $\mathbf{W}_p$ is the $p$-th transformation matrix. It has been pointed out that the clustering generalization error bound based on the Rademacher complexit is of order $\mathcal{O}(k/\sqrt{n})$ on unseen samples.

\subsection{Max-Min-Max Late Fusion}
Motivated by the superior performance of kernel fusion-based methods, such as SimpleMKKM \cite{liu2022simplemkkm}, and the low computational complexity of late fusion-based approaches, \cite{li2023late} propose integrating the kernel fusion-based min-max learning paradigm into late fusion MVC, which can be achieved through a max-min-max optimization framework as follows,
\begin{align}
&\max_{\{\mathbf{W}_p\}_{p=1}^{m}} \min_{\boldsymbol{\gamma}} \max_{\mathbf{H}} \quad \tr \left( \mathbf{H}^T \left( \sum_{p=1}^m \gamma_p^2 \mathbf{H}_p \mathbf{W}_p \right) \right) \\
&\st \quad  \mathbf{H}^T\mathbf{H} = \mathbf{I}_c, \sum_{p=1}^m \gamma_p = 1, \gamma_p \geq 0, \mathbf{W}_p^T \mathbf{W}_p = \mathbf{I}_c,\forall p,\nonumber
\end{align}
which maximizes the perturbation matrices $\{\mathbf{W}_p\}_{p=1}^{m}$, minimizes view coefficients $\boldsymbol{\gamma}$ and maximizes the clustering partition matrix $\mathbf{H}$. The generalization clustering performance is upper bounded with the order $\mathcal{O}(\sqrt{k/n})$.
 
 \section{Notations and Preliminaries}
Let $\mathbb{P}$ be an unknown distribution on $\mathcal{X}$, and let $\mathbb{S} = \{\mathbf{x}_i\}_{i=1}^n \subseteq \mathcal{X}$ be a set of $n$ i.i.d. samples drawn from $\mathbb{P}$. The empirical distribution $\mathbb{P}_n$ is defined as $\mathbb{P}_n(\{\mathbf{x}_i\}) = \frac{1}{n}$ for each $\mathbf{x}_i \in \mathbb{S}$, and $\mathbb{P}_n(\mathbf{x}) = 0$ for $\mathbf{x} \notin \mathbb{S}$. Given a kernel function $K : \mathcal{X} \times \mathcal{X} \rightarrow \mathbb{R}$, the corresponding kernel matrix is defined as $\mathbf{K} = \left[K(\mathbf{x}_i, \mathbf{x}_j)\right]_{i,j=1}^n$. Let $\lambda_1(\mathbf{K}) \geq \lambda_2(\mathbf{K}) \geq \cdots \geq \lambda_n(\mathbf{K}) \geq 0$ denote its eigenvalues. The Reproducing Kernel Hilbert Space (RKHS) $\mathcal{H}$ is defined as the completion of the span of $\{K(\mathbf{x}, \cdot) : \mathbf{x} \in \mathcal{X}\}$.

In multiple kernel K-means, a common choice is the convex combination of $m$ basic kernels:
\begin{align}
\mathcal{K}^{\text{multiple}} = \left\{ \sum_{p=1}^{m} \gamma_p^2 \mathbf{K}_{p} \, \middle| \, \sum^m_{p=1} \gamma_p = 1, \gamma_p \geq 0 \right\},
\end{align}
where $\mathbf{K}_p$ are the basic kernels.

\section{Multiple Kernel K-Means for Late Fusion}
In this section, we present to derive consensus clustering by integrating multiple embeddings \(\{\mathbf{H}_p\}_{p=1}^{m}\) within a simplified linear case of the Multiple Kernel KMeans framework. The K-means clustering algorithm applied to the $p$-th partition \(\mathbf{H}_p \in \mathbb{R}^{n \times r_p}\) can be formulated as follows:
\begin{align}\label{lkm}
\min_{\mathbf{Y} \in \textrm{Ind}} \quad & \tr(\mathbf{H}_p\mathbf{H}_p^T) - \tr(\mathbf{Y}(\mathbf{H}_p\mathbf{H}_p^T) \mathbf{Y}(\mathbf{Y}^T\mathbf{Y})^{-1}).
\end{align}
It is well established that K-means can be more effectively conducted in a kernel space, where data points are more easily separable. Given a set of orthogonal embeddings \(\{\mathbf{H}_p\}_{p=1}^{m}\), we can construct a corresponding set of linear kernels \(\{ \mathbf{K}_p = \mathbf{H}_p\mathbf{H}_p^T\}_{p=1}^{m}\). For the LFMVC task, we propose to learn the consensus clustering using a consensus kernel \(\mathbf{K}_{\boldsymbol{\gamma}}\), which is obtained through a weighted aggregation of the individual kernels: \(\mathbf{K}_{\boldsymbol{\gamma}} = \sum_{p=1}^{m}\gamma_p^2 \mathbf{K}_p\). The MKKM-based LFMVC can then be formulated as follows:
\begin{align}\label{GMLKM}
\min_{\mathbf{Y},\boldsymbol{\gamma}} \quad & \tr\left(\sum_{p=1}^{m}\gamma_p^2 \mathbf{H}_p\mathbf{H}_p^T\right) \nonumber\\
\quad &- \tr\left(\mathbf{Y}\left(\sum_{p=1}^{m}\gamma_p^2 \mathbf{H}_p\mathbf{H}_p^T\right) \mathbf{Y}(\mathbf{Y}^T\mathbf{Y})^{-1}\right) \\
\st \quad & \mathbf{Y} \in \textrm{Ind},  \sum_{p=1}^{m}\gamma_p=1, \gamma_p \geq 0,  \forall i. \nonumber
\end{align}
Although applying the MKKM framework from Eq.~\eqref{GMLKM} to the task of LFMVC may seem straightforward, several key insights deserve emphasis. First, MKKM demonstrates significant efficiency by operating in linear time when using linear kernels derived from candidate embeddings. This characteristic aligns with the inherent advantages of late fusion-based approaches, making MKKM particularly suitable for large-scale datasets. Second, by employing the local Rademacher complexity technique, we establish that the generalization error for MKKM is of order \(\mathcal{O}(1/n)\), which provides a much sharper bound than those achieved by previously discussed LFMVC methods. Furthermore, the generalization error bound can be further improved by enhancing the proportion of the principal eigenvalue.

\section{Generalization Error Analysis of MKKM}

The generalization error of the K-means algorithm is defined as the expected distance between unseen data and their corresponding cluster centers. To formalize this, we consider a class of functions to construct our hypothesis space:
\begin{align}\label{loss_function}
    \mathcal{L} =& \Bigg\{ \ell : \mathbf{x} \mapsto \min_{\mathbf{y} \in \{ \mathbf{e}_1, \ldots, \mathbf{e}_k\}} \|\Phi_{\boldsymbol{\gamma}}(\mathbf{x}) - \mathbf{C} \mathbf{y}\|_{\mathcal{H}}^2 \,\Bigg|\, \nonumber\\&|\Phi_{\boldsymbol{\gamma}}(\mathbf{x}_i)^T \Phi_{\boldsymbol{\gamma}}(\mathbf{x}_j)\| \leq b, \, \forall \mathbf{x}_i, \mathbf{x}_j \in \mathcal{X}, \, \mathbf{C} \in \mathcal{H}^k \Bigg\},
\end{align}
where $\Phi_{\boldsymbol{\gamma}}(\mathbf{x})=[\gamma_1\Phi_{1}(\mathbf{x})^T,\dots,\gamma_m\Phi_{m}(\mathbf{x})^T]^T:\mathbb{R}^d\mapsto \mathcal{H}$. Suppose $\mathbf{c}_j$ is the $j$-th clustering centroid and $f_{\mathbf{c}}=(f_{c_{1}},\ldots,f_{c_{k}})$, where $j \in \{1, \ldots, k\}$. It follows that the mapping function $f_j$ is bounded within $[0, 4b]$.
\begin{align}\label{score_function}
     f_{c_j}\left(\mathbf{x}\right) &= \|\Phi_{\boldsymbol{\gamma}}(\mathbf{x}) - \mathbf{c}_j\|_{\mathcal{H}}^2 = \left\| \Phi_{\boldsymbol{\gamma}}(\mathbf{x}) - \frac{1}{|\mathbf{c}_j|}\sum_{i \in \mathbf{c}_j} \Phi_{\boldsymbol{\gamma}}(\mathbf{x}_i)\right\|_{\mathcal{H}}^2 \nonumber\\
    &\leq 2 \|\Phi_{\boldsymbol{\gamma}}(\mathbf{x})\|^2 + \frac{2}{|\mathbf{c}_j|^2}\sum_{i_1, i_2 \in \mathbf{c}_j} \langle \Phi_{\boldsymbol{\gamma}}(\mathbf{x}_{i_1}), \Phi_{\boldsymbol{\gamma}}(\mathbf{x}_{i_2}) \rangle \nonumber\\
    &\leq 4b.
\end{align}
The generalization error $R(f)$ is defined as
\begin{align}
    R(f) := \mathbb{E}_{\mathbf{x} \sim \mathbb{P}} [\ell_f(\mathbf{x})],
\end{align}
where $\ell_f$ is the loss function. Since $\mathbb{P}$ is unknown, we estimate the empirical error using the observed data:
\begin{align}
    \hat{R}(f) := \frac{1}{n} \sum_{i=1}^n \ell_f(\mathbf{x}_i).
\end{align}

\begin{mylemma}\label{4b}
The loss function $\ell(f_{\mathbf{c}}(\mathbf{x}))=\min f_{\mathbf{c}}(\mathbf{x})$ is $1$-Lipschitz with respect to the $L_\infty$ norm, satisfying:
\begin{align}
    \ell^2 &= \min(f_1, f_2, \ldots, f_k)^2 \nonumber\\
    &\leq \max(f_1, f_2, \ldots, f_k) \min(f_1, f_2, \ldots, f_k) \nonumber\\
    &\leq 4b \min(f_1, f_2, \ldots, f_k) = 4b \ell.
\end{align}
\end{mylemma}

\begin{myproof}
Assuming that $\forall f,g\in \mathcal{H}$, $\ell(f_\mathbf{c}(\mathbf{x}))\geq\ell(g_\mathbf{c}(\mathbf{x}))$ and $\ell(g(\mathbf{x}))=g_{c_i}$, we have 
\begin{align}
    \left| \ell(f_{\mathbf{c}}(\mathbf{x})) - \ell(g_{\mathbf{c}}(\mathbf{x})) \right| &= \min f_\mathbf{c}(\mathbf{x}) - g_{c_i}(\mathbf{x}) \nonumber\\
    &\leq f_{c_i}(\mathbf{x})  - g_{c_i}(\mathbf{x})\nonumber\\
    &\leq \|f_{\mathbf{c}}(\mathbf{x})  - g_{\mathbf{c}}(\mathbf{x})\|_\infty,
\end{align}
which shows that $\ell f(\mathbf{x})$ is a $1$-Lipschitz continuous function.
\end{myproof}

Since the underlying probability distribution $\mathbb{P}$ of the data is unknown, it is not possible to directly estimate the generalization error. To address this issue, a widely adopted strategy is to combine the empirical risk with a measure of model complexity, as shown in Eq.~\eqref{R_S}. For a hypothesis $f$, the generalization error is bounded as:
\begin{align}\label{R_S}
  \mathcal{R}(f)\leq\mathcal{R}_\mathrm{emp}(f)+\sup_{f\in\mathcal{H}}\left[\mathcal{R}(f)-\mathcal{R}_\mathrm{emp}(f)\right],
\end{align}
where the second term, $\sup_{f\in\mathcal{H}}\left[\mathcal{R}(f)-\mathcal{R}_\mathrm{emp}(f)\right]$, can be controlled using the Rademacher complexity $\Re(\mathcal{H})$:
\begin{align}
\sup_{f\in\mathcal{H}}\left[\mathcal{R}(f)-\mathcal{R}_\mathrm{emp}(f)\right]\leq\Re(\mathcal{H}).
\end{align}

Compared to the global Rademacher complexity, the local Rademacher complexity \cite{LRC_2005} provides more refined bounds on the generalization error, avoiding the overly conservative estimates from global metrics. Its specific definition is given by:
\begin{align}
    \Re_n(\mathcal{H}, r) := \Re_n \left( \left\{ f \in \mathcal{H} : \mathbb{E}[f^2] \le r \right\} \right),
\end{align}
where \( r \) represents a tolerance for the complexity or error of the functions, and must satisfy \( r > 0 \).

Given the significant challenges in directly assessing local Rademacher complexity, an approximate estimation method based on the ratio of the eigenvalues to the secondary eigenvalues of the integral operator can be adopted \cite{Kernel_JMLR_2003}. Building on the concept of Principal Eigenvalue Proportion (PEP) introduced in \cite{PEP_AAAI_2017}, specifically Definitions 1 and 2, we utilize these to estimate the local Rademacher complexity of our model presented in Eq.~\eqref{GMLKM}.

\begin{mydefinition}\cite{PEP_AAAI_2017}
    Assume $K$ is a Mercer kernel with its eigenvalue decomposition given by $K\left(\mathbf{x}, \mathbf{x}^{\prime}\right)=\sum_{j=1}^{\infty} \lambda_j(K) \phi_j(\mathbf{x}) \phi_j\left(\mathbf{x}^{\prime}\right)$, where the sequence $\left(\lambda_j(K) \right)_{j=1}^{\infty}$ represents the eigenvalues of the kernel function arranged in decreasing order. The $t$-Principal Eigenvalue Proportion ($t$-PEP) of $K$, for $t \in \mathbb{N}_{+}$, is defined as follows:
\begin{align}
\beta(K, t)=\frac{\sum_{i=1}^t \lambda_i(K)}{\sum_{i=1}^{\infty} \lambda_i(K)} .    
\end{align}
\end{mydefinition}

\begin{mydefinition}\cite{PEP_AAAI_2017}
    Assume  \( K \) is a kernel function and let \( \mathbf{K} = \left[\frac{1}{n}K(x_i, x_j)\right]_{i,j=1}^n \) denote its corresponding kernel matrix. The \( t \)-empirical principal eigenvalue proportion of \( \mathbf{K} \), for \( t \in \{1, \dots, n - 1\} \), is given by
\begin{align}
\hat{\beta}(\mathbf{K}, t) = \frac{\sum_{i=1}^{t} \lambda_i(\mathbf{K})}{\text{tr}(\mathbf{K})},
\end{align}
where \( \lambda_i(\mathbf{K}) \) represents the \( i \)-th largest eigenvalue of \( \mathbf{K} \), and \( \text{tr}(\mathbf{K}) \) denotes the trace of the matrix \( \mathbf{K} \).
\end{mydefinition}

\begin{mylemma}\label{RH} Assume $\kappa=\sup_{p\in\{1,2,\ldots,m\}}$tr$(K_p) <\infty$,
the Theorem 6 of \cite{LPMKL_JMLR_2012} and the Corollary 1 of \cite{PEP_AAAI_2017} show that for $\forall K \in \mathcal{K}^{\text {multiple }}$, $D> 0$, denote
$$
\mathcal{H}_K^{\text {multiple }}=\left\{f_w:x\mapsto\left\langle\mathbf{w}, \Phi_K(\mathbf{x})\right\rangle:\|\mathbf{w}\| \leq D\right\},
$$
where $\mathcal{K}^{\text {multiple}}=\sum_{p=1}^{m}\gamma_p^2 K_p$. The local Rademacher complexity of the hypothesis space $\mathcal{H}_K^{\text{multiple}}$ is bounded by:
\begin{align}
    R_n(\mathcal{H}^{\text {multiple}}_{K}, r) \le &\sqrt{\frac{2}{n} \min_{\theta \geq 0} \left( \theta r + D^2m^{\frac{2}{z}-1} \sum_{j > \theta} \lambda_j (K) \right)}\nonumber\\
     \le &\sqrt{\frac{2}{n} \left( \theta r + D^2m^{\frac{2}{z}-1} \kappa \sum^{m}_{p=1} \frac{1}{\beta(K_p,t)}\nonumber\right)},
\end{align}
Where $z$ is the optimal norm value corresponding to the proof of Proposition 2 in
\cite{LPMKL_JMLR_2012}.
\end{mylemma}

\subsection{Generalization Bounds with PEP}
Within the purview of this section, an estimation of the generalization error margin is conducted by leveraging the ratio of the dominant eigenvalues.
\begin{mytheorem}\label{mytheorem_v1}
    Let $\ell$ defined in Eq.~\eqref{loss_function} be the multiple kernel $k$-means loss function associated with the score function $f$ defined in Eq~\eqref{score_function}. Then, $\forall \delta>0$, with probability at least $1-\delta$ over the choice of a sample $\mathbb{S}=\left\{\mathbf{x}_i\right\}_{i=1}^n$ drawn i.i.d according to $\mathbb{P}$, the following inequality holds: $\forall \theta>1$ and $\forall f \in$ $\mathcal{H}_K^{\text {multiple }}$,
\begin{align}
R(f) \leq& \frac{\theta}{\theta-1} \hat{R}(f) + \frac{c_1}{\sqrt{nz}} + \frac{c_2+c_3}{n} ,\nonumber
\end{align}
where $z=\sum^{m}_{p=1} \beta(K_p,t)$, $  c_1= 6\theta b D \sqrt{2\kappa m^{\frac{2}{z}-1}}, c_2 = 48 \theta b t,c_3=(44 b + 20 b \theta) \log \frac{1}{\delta}$.    
\end{mytheorem}

\begin{myproof}
It follows from Lemma 1 that $\ell_{f}(x)^2 \le 4b\ell_{f}(x)$, we get $\mathbb{E} \ell^2_{f}(x) \le 4b \mathbb{E} \ell_{f}(x)$. According to Theorem 3.3 of \cite{LRC_2005}, we can know with at least $1 - \delta$ probability that:
\begin{equation}\label{loacal_rademacher}
    \forall \theta > 1, R(f) \le \frac{\theta}{\theta-1} R_{\text{emp}}(f) + \frac{6 \theta r^*}{4b} + \frac{e}{n},
\nonumber\end{equation}
where $e=\log \frac{1}{\delta} (44 b + 20 b \theta)$ and $r^*$ is the fixed point of $4bR_n(\mathcal{L}, r)$. According to Lemma \ref{4b}, we get $\ell_{f}$ is a $1$-Lipschitz with respect to the $L_\infty$ norm. According to Lemma 5 of \cite{cortes2016structured}, we get $R_n(\mathcal{L}, r) \le  R_n(\mathcal{H}^{\text{multiple}}_{K}, r)$. According to Lemma \ref{RH}, we know
\[
R_n(\mathcal{H}^{\text{multiple}}_{K}, r) \le \sqrt{\frac{2}{n} \left( \theta r + D^2 m^{\frac{2}{z}-1}\kappa \sum^{m}_{p=1} \frac{1}{\beta(K_p,t)} \right)}.
\]
Then we have
\begin{align*}
    4b\mathcal{R}_n(\mathcal{L}, r) \leq& 4b \mathcal{R}_n
    (\mathcal{H}^{\text{multiple}}_{K}, r) \\
    \leq& 4b\sqrt{\frac{2}{n} \left( t \cdot r + D^2m^{\frac{2}{z}-1} \kappa \sum^{m}_{p=1} \frac{1}{\beta(K_p,t)} \right)}.
\end{align*}
Let $a = \frac{32b^2t}{n}$, $c=\frac{32 \kappa b^2 D^2m^{\frac{2}{z}-1}}{n} \sum^{m}_{p=1} \frac{1}{\beta(K_p,t)}$, and by solving $r^2 - ar -c = 0$, we can obtain:
\begin{align*}
    r^{*} = &\frac{a + \sqrt{a^2+4c}}{2} \leq \frac{a + \sqrt{a^2}+ \sqrt{4c}}{2}= a + \sqrt{c} \\ =& \frac{32 b^2 t}{n} + \sqrt{\frac{32 \kappa b^2 D^2m^{\frac{2}{z}-1}}{n} \sum^{m}_{p=1} \frac{1}{\beta(K_p,t)} }
\end{align*}
then bring $r^*$ back to Eq.~\eqref{loacal_rademacher} and complete this proof.
\end{myproof}
\begin{myassumption}\label{beta=1}
 Assume that the kernel function spectrum satisfies the polynomial spectrum decay rate, that is 

$$\exists \gamma > 1 : \lambda_i(K) = O(i^{-\gamma}).$$
The polynomial spectral decay rate is a common assumption, which is satisfied by both finite-rank kernels and convolution kernels. Assuming that the kernel function \( K \) satisfies the polynomial spectral decay and we assume that the t-PEP value corresponding to the $v$-th kernel is the smallest, we can get
\begin{align}
\sum_{i=t+1}^{\infty} \lambda_i^{v}(K) &= \sum_{i=t+1}^{\infty} O(i^{-\gamma})  \leq \int_t^\infty O(x^{-\gamma}) \, dx \nonumber \\&= O\left(\frac{x^{1-\gamma}}{1-\gamma}\Big|_t^\infty\right)   = O\left(\frac{t^{1-\gamma}}{\gamma-1}\right).
\end{align}
Therefore, there is
$$
\sum_{i=1}^{t} \lambda_i^{v}(K) = \tr(K_v) - \sum_{i=t+1}^{\infty} \lambda_i^{v}(K) \geq \tr(K_v) - O\left(\frac{t^{1-\gamma}}{\gamma-1}\right),
$$
$$
\frac{\sum_{i=1}^{t} \lambda_i^{v}(K)}{\sum_{i=1}^{\infty} \lambda_i^{v}(K)}\geq \Omega(\left(\frac{t^{1-\gamma}}{\gamma-1}\right)).
$$
In this case, the generalization error bound converges to
$$
\mathcal{O}\left(\frac{1}{\sqrt{n\sum^{m}_{p=1} \beta(K_p,t)}} + \frac{1}{n}\right).
$$
If $\ t^{\gamma-1} \left(\gamma-1\right) \geq \frac{\sqrt{n}}{m}$  means  $ t \geq \left(\frac{\sqrt{n}}{m\left(\gamma-1\right)}\right)^{\frac{1}{\gamma-1}} $, the convergence rate can reach $ \mathcal{O}\left(\frac{1}{n}\right) $.
\end{myassumption}

\begin{myremark}   
 $\gamma_p$ represents the weight for each kernel function $K_p$, while PEP measures the complexity of the kernel functions. When the proportion of the principal eigenvalue is low, \(\beta(K_p, t)\) is also low, indicating a higher complexity of the kernel function. This conclusion further supports the strategy in multiple kernel learning of assigning appropriate weights to kernel functions of varying complexities to achieve models with better generalization performance.
\end{myremark}


\subsection{Generalization Bounds with Empirical PEP}
Since it is not easy to compute the value of PEP, we have to
use the empirical PEP for practical kernel learning. Fortunately, we can also use it to derive sharp bound.
\begin{mytheorem}
Let $\ell$ defined in Eq.~\eqref{loss_function} be the multiple kernel $k$-means loss function associated with the score function $f$ defined in Eq.~\eqref{score_function}. Then, with $1-\delta, \forall \theta \geq 1, f \in \mathcal{H}_K^{\text {multiple }}$, we have
\begin{align}
R(f) \leq \frac{\theta}{\theta-1} \hat{R}(f) + \frac{c_3}{\sqrt{n\hat{z}}}+ \frac{c_4+c_5}{n}
\end{align}
where $\hat{z}=\sum^{m}_{p=1} \hat{\beta}(\mathbf{K}_p,t), c_3=12 \theta D \sqrt{2 \kappa m^{\frac{2}{z}-1}}, c_4=384\theta b t, c_5=(11+20 b \theta+\frac{39\theta}{b}) \log (\frac{3}{\delta}).$ \\
\begin{myproof}
To prove the  theorem, we introduce the empirical localized Rademacher complexity of $\mathcal{H}$, which is defined as:
\[
\hat{R}_n(\mathcal{H}, r) := R_n \left( \left\{ f \in \mathcal{H} : \frac{1}{n} \sum^n_{i=1}(f(\mathbf{x}_i))^2 \le r \right\} \right).
\]From \cite{LRC_2005}, it is known that with at least probability \(1 - \delta\), for all \(\theta > 1\), the following formula holds:
\begin{align} \label{data_local_rademacher}
    R(f) \leq \frac{\theta}{\theta - 1} R_{\text{emp}}(f) + \frac{6\theta \hat{r}^*}{4b} + \frac{\log\left(\frac{3}{\delta}\right)(44b + 20 b \theta)}{n},
\end{align}
where \(\hat{r}^*\) is the fixed point of $ 8b \hat{R}_n(\mathcal{L}, 2r) + \frac{13 \log(3/\delta)}{n}$
 and the empirical localized Rademacher complexity of $\mathcal{H}$ is defined as:
\[
\hat{R}_n(\mathcal{H}, r) := R_n \left( \left\{ f \in \mathcal{H} : \frac{1}{n} \sum^n_{i=1}(f(\mathbf{x}_i))^2 \le r \right\} \right).
\]
According to the Lemma \ref{RH}, we have
\begin{align*}
    &8b \hat{R}_n(\mathcal{L}, 2r) + \frac{13 \log(3/\delta)}{n} \\ \leq& 8b \hat{R}_n
    (\mathcal{H}^{\text{multiple}}_{K}, 2r) 
    + \frac{13 \log(3/\delta)}{n} \\
    \leq& 8b\sqrt{\frac{2}{n} \left( t \cdot 2 r + D^2 m^{\frac{2}{z}-1} \kappa \sum^{m}_{p=1} \frac{1}{\hat{\beta}(\mathbf{K}_p,t)} \right)} \\&+ \frac{13 \log(3/\delta)}{n}.
\end{align*}
Similar to the proof of the Theorem \ref{mytheorem_v1}, we need to solve for
\begin{align}
    r=&8b \sqrt{\frac{2}{n}\left(t \cdot 2r + D^2m^{\frac{2}{z}-1} \kappa \sum^{m}_{p=1} \frac{1}{\hat{\beta}(\mathbf{K}_p,t)} \right)} \nonumber\\&+ \frac{13 \log\left(\frac{3}{\delta}\right)}{n} \nonumber
\end{align}
Defining $a = \frac{256 t b^2}{n},c = \frac{128 b^2 D^2 m^{\frac{2}{z}-1} \kappa}{n}\sum^{m}_{p=1} \frac{1}{\hat{\beta}(\mathbf{K}_p,t)}$ and $d = \frac{13 \log\left(\frac{3}{\delta}\right)}{n}.$
The above is equivalent to solving
\[
r^2 - (2d + a)r - (c - d^2) = 0
\]
for a positive root. It is easy to verify that
\begin{align*}
    \hat{r}^* \leq& \frac{2d + a + \sqrt{a^2 + 4da + 4c}}{2} \\
    \leq& \frac{2d + a + \left(a + 2d + 2\sqrt{c}\right)}{2} \\
    =& a + 2d + \sqrt{c}  = \frac{256 t b^2}{n } + \frac{26 \log\left(\frac{3}{D}\right)}{n} \\&+ 8 b D  \sqrt{\frac{  2\kappa m^{\frac{2}{z}-1} \sum^{m}_{p=1} \frac{1}{\hat{\beta}(\mathbf{K}_p,t)}}{n}}  
\end{align*}
Then we bring $\hat{r}^*$ back to Eq.~\eqref{data_local_rademacher} and complete this proof. Same as the assumption \ref{beta=1}, when  $\ t^{\gamma-1} \left(\gamma-1\right) \geq \frac{\sqrt{n}}{m}$ means $t \geq \left(\frac{\sqrt{n}}{m\left(\gamma-1\right)}\right)^{\frac{1}{\gamma-1}}$, the convergence rate can reach $\mathcal{O}\left(\frac{1}{n}\right)$.
\end{myproof}
\end{mytheorem}

\section{Graph Filter Enhanced Multiple Linear K-means}
In this section, we introduce a strategy within the linear case of the MKKM framework to denoise base partitions by leveraging smooth representations obtained through an optimal graph filter, which aims to enhance the principal eigenvalue proportions.

In graph signal processing, natural signals are expected to be smooth across adjacent nodes, aligning with the underlying graph structure. Smoother signals often lead to clearer clustering, consistent with the cluster and manifold assumption \cite{ACMK_TKDE_2023}. To achieve this, it is crucial to smooth raw features, reducing high-frequency noise while preserving key graph-based properties. 

We start by constructing a set of affinity graphs $\{\mathbf{A}_p\}_{p=1}^{m}$, with each $\mathbf{A}_p \in \mathbb{R}^{n \times n}$. For the \(p\)-th data view, the normalized adjacency matrix is \(\mathbf{D}_{p}^{-\frac{1}{2}} \mathbf{A}_{p} \mathbf{D}_{p}^{-\frac{1}{2}}\), where \(\mathbf{D}_{p}\) is the diagonal degree matrix. The corresponding normalized graph Laplacian is:
\begin{align}
\mathbf{L}_p = \mathbf{I} - \mathbf{D}_{p}^{-\frac{1}{2}} \mathbf{A}_{p} \mathbf{D}_{p}^{-\frac{1}{2}}.
\end{align}
A common technique is to apply a low-pass graph filter, typically expressed as:
\begin{align}
\hat{\mathbf{x}} = \sum_{o=1}^{\bar{o}} \left(\frac{\mathbf{I} + \mathbf{D}_{p}^{-\frac{1}{2}} \mathbf{A}_{p} \mathbf{D}_{p}^{-\frac{1}{2}}}{2}\right)^o \mathbf{x} = \mathbf{G}_p \mathbf{x},
\end{align}
where \(o\) represents the \(o\)-hop neighborhood.
\begin{algorithm}[htbp]
	\caption{Algorithm for the problem in Eq.~\eqref{gGMLKM}.}
	\label{alg_gGMLKM}
	\begin{algorithmic}[1] 
		\REQUIRE{Base partitions $\{\mathbf{H}_p\}_{p=1}^{m}$, the cluster number $k$, the neighborhood size is set to 5.}
		\STATE{Compute the probabilistic affinity graph $\{\mathbf{S}_r\}_{r=1}^{m}$ according to \cite{can};}
		\STATE{Initialize $\boldsymbol{\gamma},\boldsymbol{\mu}$,$\mathbf{Y}$;}
		\REPEAT
		\STATE{Update $\mathbf{Y}$ by CD algorithm \cite{CDKM_2022};}
		\STATE{Update $\boldsymbol{\gamma}$ according to Eq.~\eqref{update_gamma};}
        \STATE{Update $\boldsymbol{\mu}$ according to Eq.~\eqref{opt_mu};}
		\UNTIL{Converges}
		\ENSURE{The discrete clustering result $\mathbf{Y}$.}
	\end{algorithmic}
\end{algorithm}

Traditional methods, such as those in \cite{GC_TKDE_2023}, typically use fixed pre-defined graph filters, which may not adapt well to complex multi-view data. To address this limitation, we propose learning an optimal low-pass graph filter to enhance smoothness across embeddings. This is achieved by expressing the smoother embedding \(\hat{\mathbf{H}}_p\) as:  
\begin{align}\label{GH}
\hat{\mathbf{H}}_p = \left(\sum_{i=1}^{m}\mu_i \mathbf{G}_i\right) \mathbf{H}_p, 
\end{align}  
where \(\{\mu_i\}_{i=1}^{m}\) are learnable combination coefficients, and \(\{\mathbf{G}_i\}\) are graph-based filters associated with different views. This formulation ensures that the derived embedding is not only smoother but also incorporates complementary information from multiple views, optimizing the clustering process. Based on this, we propose the Graph filter-enhanced Multiple Linear K-Means (GMLKM):
\begin{align}\label{gGMLKM}
\min \quad &\tr\left(\sum_{p=1}^{m}\gamma_p^2 \tilde{\mathbf{G}}\mathbf{H}_p\mathbf{H}_p^T\tilde{\mathbf{G}}\right) \\
-& \tr\left(\mathbf{Y}^T(\sum_{p=1}^{m}\gamma_p^2 \tilde{\mathbf{G}}\mathbf{H}_p\mathbf{H}_p^T\tilde{\mathbf{G}}) \mathbf{Y}(\mathbf{Y}^T\mathbf{Y})^{-1}\right)   \nonumber \\
\st \quad &  \tilde{\mathbf{G}} = \sum_{i=1}^{m} \mu_i \mathbf{G}_i, \mathbf{Y} \in \textrm{Ind}, \nonumber \\
&\sum_{p=1}^{m}\gamma_p=1,\sum_{i=1}^{m}\mu_i=1,\gamma_p \geq 0, \mu_i \geq 0,\forall p,i. \nonumber
\end{align} 
This filtering approach is preferred as it captures high-order relationships while maintaining linear computational complexity through sparse matrix multiplication. 

\begin{myremark}
PEP quantifies the proportion of variance captured by the leading eigenvalues of the kernel matrix. Higher PEP values suggest that most of the variance is explained by the dominant
eigenvalues,enabling effective clustering as the data can be well-represented in a lower-dimensional space.The adaptive graph filter enhances PEP by increasing the smoothness and
relevance of the data representation,resulting in a more concentrated and informative eigenvalue distribution.
\end{myremark}

\subsection{Alternate Optimization}
The objective in Eq.~\eqref{gGMLKM} involves optimizing four variables. We propose a three-step alternating optimization procedure by optimizing one variable while keeping the others fixed.

\textbf{Optimization $\mathbf{Y}$:}  With \(\boldsymbol{\gamma}\) and \(\boldsymbol{\mu}\) fixed, the optimization problem in Eq.~\eqref{gGMLKM} with respect to \(\mathbf{Y}\) reduces to:
\begin{align}
    \max_{\mathbf{Y} \in \textrm{Ind}} \quad \tr\left(\mathbf{Y}^T(\tilde{\mathbf{H}}\tilde{\mathbf{H}}^T) \mathbf{Y}(\mathbf{Y}^T\mathbf{Y})^{-1}\right),
\end{align}
where \(\tilde{\mathbf{H}} =  \tilde{\mathbf{G}}[\gamma_1 \mathbf{H}_1,\gamma_2 \mathbf{H}_2,\ldots,\gamma_m \mathbf{H}_m]\). This formulation resembles the K-means algorithm applied to \(\Tilde{\mathbf{H}}\). Hence, the optimal \(\mathbf{Y}\) can be efficiently obtained using the Coordinate Descent (CD) method \cite{CDKM_2022}.

\textbf{Optimization $\boldsymbol{\gamma}$:}  With \(\mathbf{Y}\) and \(\boldsymbol{\mu}\) fixed, the optimization problem in Eq.~\eqref{gGMLKM} with respect to \(\boldsymbol{\gamma}\) simplifies to:
\begin{align}
\min\; \;\sum_{p=1}^{m} \gamma_p^2 \alpha_p, \;\;\st \;\;\sum_{p=1}^{m}\gamma_p=1, \gamma_p \geq 0,  \forall p, 
\end{align}
where \(\alpha_p =-\tr(\mathbf{Y}^T \tilde{\mathbf{G}}\mathbf{H}_p\mathbf{H}_p^T\tilde{\mathbf{G}}\mathbf{Y}(\mathbf{Y}^T\mathbf{Y})^{-1})+\tr( \tilde{\mathbf{G}}\mathbf{H}_p\mathbf{H}_p^T\tilde{\mathbf{G}}) \). The optimal solution \cite{rmkkm_ijcai_2015} for \(\gamma_p\) is given by:
\begin{align}\label{update_gamma}
  \gamma_p = \frac{\frac{1}{\alpha_p}}{\sum_{p'=1}^{m}\frac{1}{\alpha_{p'}}}.
\end{align}

\textbf{Optimization $\boldsymbol{\mu}$:}  With \(\mathbf{Y}\) and \(\boldsymbol{\gamma}\) fixed, the optimization problem in Eq.~\eqref{gGMLKM} with respect to \(\boldsymbol{\mu}\) is reformulated as:
\begin{align}\label{opt_mu}
\min \;\; \boldsymbol{\mu}^T \mathbf{M} \boldsymbol{\mu}, \;\; \st \;\; \sum_{i=1}^{m}\mu_i=1, \mu_i \geq 0, \forall i,   
\end{align}
where the matrix \(\mathbf{M} \in \mathbb{R}^{m \times m}\) is defined as \(\mathbf{M}_{ij} =-\text{tr}(\mathbf{Y}^T  \mathbf{G}_i(\sum_{p=1}^{m}\gamma_p \mathbf{H}_p\mathbf{H}_p^T) \mathbf{G}_j \mathbf{Y}(\mathbf{Y}^T\mathbf{Y})^{-1})+\text{tr}( \mathbf{G}_i(\sum_{p=1}^{m}\gamma_p\mathbf{H}_p\mathbf{H}_p^T) \mathbf{G}_j) \). This is a standard quadratic programming problem. The whole optimization procedure in solving Eq.~\eqref{gGMLKM} is outlined in Algorithm \ref{alg_gGMLKM}.
\subsection{Convergence Analysis}
The optimization problem in Eq.~\eqref{gGMLKM} is evidently lower-bounded, and the variable updates for updating $\mathbf{Y}$, $\boldsymbol{\gamma}$, and $\boldsymbol{\mu}$ monotonically decrease the objective function. Thus, convergence of Algorithm \ref{alg_gGMLKM} is guaranteed.

\section{Experiments}
\subsection{Datasets}
In Table \ref{tab:datasets}, we evaluate our method on ten benchmark datasets: Lung (D1), JAFFE (D2), CSTR (D3), ISOLET(D4), WAP (D5), Mfeat (D6), K1b (D7), MouseBladder (D8), PBMC (D9), and CBMC (D10). These datasets cover a range of domains, including image datasets (JAFFE, ISOLET, Mfeat), biological datasets (Lung, MouseBladder, PBMC, CBMC), and text corpora (CSTR, WAP, K1b). 
\begin{table}[h]
	\centering
	\caption{Summary of ten MVC datasets}
	\label{tab:datasets}
	\begin{tabular}{@{}c@{\hspace{15pt}}ccccc@{}}
		\toprule
		\textbf{ID} & \textbf{Dataset} &  \textbf{Samples} &  \textbf{Features} &  \textbf{Classes} \\ \midrule
		D1 & Lung    & 203   & 3312   & 5     \\ 
        D2 & JAFFE    & 213   & 676   & 10     \\ 		
        D3 & CSTR      & 476   & 1000    & 4    \\ 
		D4 & ISOLET      & 1560   & 617    & 26 \\ 
		D5 & WAP      & 1560   & 8460    & 20  \\ 
		D6 & Mfeat        & 2000   & 240   & 10   \\
		D7 & K1b & 2340  & 21839     & 6    \\ 
        D8 & MouseBladder   & 2746   & 11829    & 16  \\ 
        D9 & PBMC     & 4271 & 12206     & 8    \\
        D10 & CBMC     & 8617 & 1703     & 15    \\
		\bottomrule
	\end{tabular}     
\end{table}

\begin{table*}[ht]
\caption{Clustering results in terms of ACC/NMI/ARI}
\label{tab:performance}.
\resizebox{0.98\textwidth}{!}{
    \setlength{\tabcolsep}{6pt}
\begin{tabular}{ccccccccccc} \\ 
\toprule
 & D1    & D2    & D3    & D4    & D5    & D6    & D7  & D8  & D9  & D10  \\ \midrule
\multicolumn{11}{c}{ACC($\%$)} \\ \midrule  
avgH  &  79.5 $\pm$ 7.3 & 87.4 $\pm$ 7.9 & 76.5 $\pm$ 14.9 & 55.4 $\pm$ 2.9 & 40.3 $\pm$ 6.2 & 85.0 $\pm$ 5.6 & 81.2 $\pm$ 15.8 & 53.7 $\pm$ 3.7 & 58.8 $\pm$ 2.1 & 62.1 $\pm$ 3.1 \\
AWP  &  64.0 $\pm$ 0.0 & 62.0 $\pm$ 0.0 & 66.7 $\pm$ 0.0 & 45.2 $\pm$ 0.0 & 33.7 $\pm$ 0.0 & 64.2 $\pm$ 0.0 & 85.4 $\pm$ 0.0 & 53.4 $\pm$ 0.0 & 53.7 $\pm$ 0.0 & 59.7 $\pm$ 0.0 \\
LFMVC  &  81.5 $\pm$ 7.1 & 89.7 $\pm$ 7.2 & 80.8 $\pm$ 7.0 & 55.7 $\pm$ 2.7 & 42.2 $\pm$ 3.2 & 84.6 $\pm$ 9.0 & 83.4 $\pm$ 8.5 & 54.4 $\pm$ 2.2 & 59.4 $\pm$ 2.3 & 62.2 $\pm$ 2.9 \\
OPLF  &  86.7 $\pm$ 0.0 & 97.3 $\pm$ 2.7 & 90.3 $\pm$ 0.0 & 59.3 $\pm$ 1.1 & 47.2 $\pm$ 1.4 & 94.9 $\pm$ 2.8 & 92.0 $\pm$ 0.3 & 61.6 $\pm$ 0.5 & 64.1 $\pm$ 0.7 & 66.0 $\pm$ 0.2 \\
LFLKA  &  81.8 $\pm$ 9.1 & 89.5 $\pm$ 7.2 & 80.8 $\pm$ 7.1 & 55.7 $\pm$ 2.7 & 42.9 $\pm$ 3.4 & 85.8 $\pm$ 5.2 & 86.0 $\pm$ 8.0 & 54.8 $\pm$ 2.4 & 59.1 $\pm$ 2.0 & 62.9 $\pm$ 2.6 \\
ALMVC  &  79.0 $\pm$ 12.1 & 83.5 $\pm$ 9.9 & 80.6 $\pm$ 7.0 & 56.8 $\pm$ 3.4 & 41.9 $\pm$ 3.6 & 84.6 $\pm$ 9.0 & 83.1 $\pm$ 8.5 & 54.1 $\pm$ 2.3 & 58.9 $\pm$ 1.8 & 62.2 $\pm$ 3.0 \\
M3LF  &  79.7 $\pm$ 7.1 & 84.2 $\pm$ 7.9 & 76.1 $\pm$ 12.6 & 56.3 $\pm$ 2.7 & 42.0 $\pm$ 3.1 & 82.7 $\pm$ 8.3 & 82.1 $\pm$ 4.5 & 53.8 $\pm$ 2.7 & 60.0 $\pm$ 2.6 & 62.2 $\pm$ 2.6 \\
sLGm  &  79.6 $\pm$ 7.3 & 84.2 $\pm$ 7.4 & 80.7 $\pm$ 7.1 & 55.8 $\pm$ 2.7 & 41.6 $\pm$ 2.7 & 84.6 $\pm$ 9.0 & 84.0 $\pm$ 8.5 & 54.6 $\pm$ 2.0 & 59.0 $\pm$ 2.1 & 62.2 $\pm$ 2.9 \\
RIWLF  &  79.5 $\pm$ 8.0 & 86.6 $\pm$ 8.2 & 80.6 $\pm$ 7.0 & 57.1 $\pm$ 2.6 & 42.2 $\pm$ 2.7 & 84.6 $\pm$ 9.0 & 84.0 $\pm$ 8.5 & 54.5 $\pm$ 2.4 & 59.0 $\pm$ 2.1 & 62.2 $\pm$ 3.0 \\
GMLKM  &  87.2 $\pm$ 0.0 & 97.3 $\pm$ 2.7 & 91.8 $\pm$ 0.0 & 60.4 $\pm$ 0.6 & 47.7 $\pm$ 0.5 & 94.7 $\pm$ 2.7 & 93.0 $\pm$ 0.0 & 62.3 $\pm$ 0.5 & 66.6 $\pm$ 0.6 & 68.5 $\pm$ 0.3 \\
\midrule
\multicolumn{11}{c}{NMI($\%$)} \\ \midrule  
avgH  &  56.3 $\pm$ 6.8 & 91.8 $\pm$ 4.1 & 64.8 $\pm$ 18.6 & 73.9 $\pm$ 1.6 & 48.2 $\pm$ 6.6 & 86.4 $\pm$ 2.9 & 68.4 $\pm$ 18.7 & 61.7 $\pm$ 1.8 & 59.1 $\pm$ 1.5 & 67.5 $\pm$ 0.9 \\
AWP  &  42.3 $\pm$ 0.0 & 69.8 $\pm$ 0.0 & 47.4 $\pm$ 0.0 & 65.0 $\pm$ 0.0 & 42.6 $\pm$ 0.0 & 68.7 $\pm$ 0.0 & 66.9 $\pm$ 0.0 & 61.6 $\pm$ 0.0 & 53.1 $\pm$ 0.0 & 65.8 $\pm$ 0.0 \\
LFMVC  &  59.5 $\pm$ 7.4 & 92.7 $\pm$ 3.8 & 69.5 $\pm$ 4.0 & 74.2 $\pm$ 1.3 & 50.3 $\pm$ 1.3 & 86.3 $\pm$ 4.2 & 70.8 $\pm$ 7.8 & 61.6 $\pm$ 1.7 & 60.5 $\pm$ 1.3 & 67.8 $\pm$ 0.5 \\
OPLF  &  66.6 $\pm$ 0.0 & 96.8 $\pm$ 1.7 & 75.3 $\pm$ 0.0 & 77.0 $\pm$ 0.5 & 52.0 $\pm$ 0.2 & 90.7 $\pm$ 1.8 & 80.7 $\pm$ 1.2 & 68.6 $\pm$ 0.1 & 61.7 $\pm$ 0.6 & 68.2 $\pm$ 0.3 \\
LFLKA  &  60.5 $\pm$ 9.3 & 92.7 $\pm$ 3.8 & 69.6 $\pm$ 3.9 & 74.2 $\pm$ 1.3 & 50.5 $\pm$ 1.0 & 86.6 $\pm$ 2.8 & 73.9 $\pm$ 7.2 & 61.9 $\pm$ 1.9 & 60.5 $\pm$ 1.4 & 67.5 $\pm$ 0.6 \\
ALMVC  &  59.4 $\pm$ 9.1 & 88.4 $\pm$ 5.1 & 69.2 $\pm$ 3.8 & 74.5 $\pm$ 1.3 & 50.0 $\pm$ 1.2 & 86.3 $\pm$ 4.2 & 70.9 $\pm$ 7.5 & 61.3 $\pm$ 1.9 & 60.1 $\pm$ 1.6 & 67.6 $\pm$ 0.7 \\
M3LF  &  56.3 $\pm$ 6.8 & 89.7 $\pm$ 4.5 & 66.5 $\pm$ 8.2 & 74.6 $\pm$ 1.4 & 50.0 $\pm$ 1.0 & 85.1 $\pm$ 3.9 & 65.4 $\pm$ 4.1 & 61.4 $\pm$ 2.1 & 59.5 $\pm$ 1.7 & 67.0 $\pm$ 0.7 \\
sLGm  &  56.6 $\pm$ 7.2 & 90.2 $\pm$ 3.8 & 69.4 $\pm$ 4.2 & 74.2 $\pm$ 1.3 & 50.2 $\pm$ 0.9 & 86.3 $\pm$ 4.2 & 72.6 $\pm$ 8.1 & 61.8 $\pm$ 1.8 & 59.4 $\pm$ 1.3 & 67.9 $\pm$ 0.3 \\
RIWLF  &  56.6 $\pm$ 7.4 & 91.0 $\pm$ 4.5 & 69.2 $\pm$ 3.8 & 74.7 $\pm$ 1.3 & 50.7 $\pm$ 1.0 & 86.3 $\pm$ 4.2 & 72.6 $\pm$ 8.0 & 61.3 $\pm$ 2.1 & 59.4 $\pm$ 1.3 & 67.6 $\pm$ 0.6 \\
GMLKM  &  66.7 $\pm$ 0.0 & 96.9 $\pm$ 1.3 & 79.4 $\pm$ 0.0 & 77.0 $\pm$ 0.4 & 47.9 $\pm$ 0.8 & 90.7 $\pm$ 1.8 & 82.7 $\pm$ 0.0 & 68.9 $\pm$ 0.5 & 60.1 $\pm$ 0.6 & 67.7 $\pm$ 0.3 \\
\midrule
\multicolumn{11}{c}{ARI($\%$)} \\
\midrule
avgH  &  59.9 $\pm$ 10.8 & 85.0 $\pm$ 8.4 & 64.2 $\pm$ 20.8 & 50.4 $\pm$ 2.6 & 31.8 $\pm$ 10.6 & 81.5 $\pm$ 5.4 & 71.1 $\pm$ 21.3 & 40.9 $\pm$ 3.3 & 45.8 $\pm$ 1.8 & 51.0 $\pm$ 2.7 \\
AWP  &  31.9 $\pm$ 0.0 & 54.1 $\pm$ 0.0 & 35.5 $\pm$ 0.0 & 38.5 $\pm$ 0.0 & 21.3 $\pm$ 0.0 & 58.0 $\pm$ 0.0 & 70.9 $\pm$ 0.0 & 41.0 $\pm$ 0.0 & 35.7 $\pm$ 0.0 & 51.8 $\pm$ 0.0 \\
LFMVC  &  65.4 $\pm$ 11.9 & 87.0 $\pm$ 8.0 & 69.0 $\pm$ 6.7 & 52.3 $\pm$ 2.6 & 33.7 $\pm$ 3.9 & 81.6 $\pm$ 7.3 & 71.7 $\pm$ 13.2 & 41.2 $\pm$ 2.3 & 46.8 $\pm$ 2.3 & 51.2 $\pm$ 2.5 \\
OPLF  &  73.6 $\pm$ 0.0 & 95.0 $\pm$ 2.9 & 80.2 $\pm$ 0.0 & 55.3 $\pm$ 0.6 & 40.2 $\pm$ 1.7 & 90.1 $\pm$ 2.7 & 86.5 $\pm$ 0.6 & 44.1 $\pm$ 0.3 & 50.4 $\pm$ 0.1 & 54.8 $\pm$ 0.1 \\
LFLKA  &  67.4 $\pm$ 14.5 & 87.0 $\pm$ 8.0 & 69.0 $\pm$ 6.7 & 52.3 $\pm$ 2.6 & 34.9 $\pm$ 4.5 & 82.3 $\pm$ 4.8 & 76.2 $\pm$ 12.6 & 41.4 $\pm$ 2.6 & 46.5 $\pm$ 2.4 & 51.7 $\pm$ 2.5 \\
ALMVC  &  63.3 $\pm$ 16.2 & 78.9 $\pm$ 9.4 & 68.6 $\pm$ 6.6 & 51.2 $\pm$ 2.3 & 33.4 $\pm$ 4.5 & 81.6 $\pm$ 7.3 & 71.3 $\pm$ 12.8 & 40.9 $\pm$ 2.5 & 46.2 $\pm$ 2.3 & 51.1 $\pm$ 2.6 \\
M3LF  &  61.7 $\pm$ 11.2 & 81.3 $\pm$ 9.1 & 64.2 $\pm$ 14.0 & 51.6 $\pm$ 1.8 & 34.8 $\pm$ 4.3 & 79.8 $\pm$ 6.7 & 69.3 $\pm$ 8.1 & 41.0 $\pm$ 2.8 & 47.3 $\pm$ 1.9 & 51.3 $\pm$ 2.4 \\
sLGm  &  60.1 $\pm$ 11.6 & 81.6 $\pm$ 8.2 & 68.9 $\pm$ 6.9 & 52.3 $\pm$ 2.6 & 33.4 $\pm$ 4.0 & 81.6 $\pm$ 7.3 & 73.6 $\pm$ 13.4 & 41.3 $\pm$ 2.5 & 46.2 $\pm$ 1.8 & 51.2 $\pm$ 2.5 \\
RIWLF  &  60.5 $\pm$ 11.7 & 83.8 $\pm$ 9.3 & 68.6 $\pm$ 6.6 & 51.5 $\pm$ 2.0 & 34.3 $\pm$ 3.7 & 81.6 $\pm$ 7.3 & 73.7 $\pm$ 13.5 & 40.8 $\pm$ 2.7 & 46.2 $\pm$ 1.8 & 51.1 $\pm$ 2.6 \\
GMLKM  &  74.7 $\pm$ 0.0 & 95.2 $\pm$ 2.4 & 84.2 $\pm$ 0.0 & 54.5 $\pm$ 0.7 & 35.2 $\pm$ 0.7 & 89.7 $\pm$ 2.8 & 89.4 $\pm$ 0.0 & 43.8 $\pm$ 0.1 & 48.2 $\pm$ 0.1 & 56.0 $\pm$ 0.1 \\
\bottomrule
\end{tabular}
}
\end{table*}

\subsection{Experimental Setup and Results}
We evaluated the proposed GMLKM method against recently developed LFMVC approaches, including AWP \cite{nie2018multiview}, LFMVC \cite{wang2019multi}, OPLF \cite{liu2021one}, LFLKA \cite{lflka_tmm_2023}, ALMVC \cite{zhang2022efficient}, M3LF \cite{M3LF_TNNLS_2024}, sLGm \cite{sLGm_MM_2024}, and RIWLF \cite{RIWLF}. Additionally, the average performance of a single embedding, denoted as AvgH, was reported as a baseline. 

Experimental settings followed those in \cite{sLGm_MM_2024}, where 12 shifted Laplacian kernels were generated using kernel parameters specified in \cite{rmkkm_ijcai_2015}. Kernel K-means was employed to derive base partitions, and the dimensionality for all methods was selected from \(\{c, 2c, 3c, 4c\}\), as recommended in \cite{wang2019multi,zhang2022efficient}. Baseline algorithms were executed using their recommended parameter settings. To mitigate the influence of randomness, each method was executed 10 times, and the average performance was reported.

For the proposed GMLKM method, the neighborhood size in Eq.~\eqref{gGMLKM} was fixed at \(5\) across all datasets without parameter tuning, while the order \(\bar{o}\) was chosen from \(\{1, 2, 3\}\). The evaluation metrics included Accuracy (ACC), Normalized Mutual Information (NMI), and Adjusted Rand Index (ARI), and the results are summarized in Table~\ref{tab:performance}.

The results demonstrate that the proposed method consistently outperforms other approaches in terms of ACC, NMI, and ARI across a variety of datasets.

\section{Conclusion}
This work presents a novel analysis using the principal eigenvalue proportion for Multiple Kernel K-means achieving a generalization error bound of \( O(1/n) \). By incorporating low pass graph filtering techniques it is expected that the principal eigenvalue proportion could be improved during for multiple kernel clustering task. We take the graph filter enhanced multiple linear K-means on base partitions in the late fusion scenarios to demonstrate the improved results. In future, we plan to conduct a thorough investigation into the theoretical and practical implications of enhancing the principal eigenvalue proportion using graph filters. 


\section{Acknowledgments}
This work was supported by the National Natural Science Foundation of China (62376146,62176001), the Shanxi Province Central Guidance for Local Science and Technology Development Special Project (YDZJSX20231D003).

\bibliography{aaai25}

\end{document}